\documentclass[letterpaper]{article} 
\usepackage[submission]{aaai23}  
\usepackage{times}  
\usepackage{helvet}  
\usepackage{courier}  
\usepackage[hyphens]{url}  
\usepackage{graphicx} 
\usepackage{natbib}  
\usepackage{caption} 
\usepackage{algorithm}
\usepackage{algorithmic}
\usepackage{epsfig}
\usepackage{amsmath}
\usepackage{amssymb}
\usepackage{bm}
\usepackage{subfigure}
\usepackage{booktabs}
\usepackage{multirow}
\usepackage{booktabs}
\usepackage{caption}
\usepackage{float}
\usepackage{algorithm, algorithmic}
\usepackage{bbding}
\usepackage{array}
\usepackage{mwe}
\usepackage{times}
\usepackage{epsfig}
\usepackage{graphicx}
\usepackage{amsmath}
\usepackage{amssymb}
\usepackage{color,xcolor}
\usepackage{float}
\usepackage{stfloats}

\usepackage{newfloat}
\usepackage{listings}
\DeclareCaptionStyle{ruled}{labelfont=normalfont,labelsep=colon,strut=off} 
\floatstyle{ruled}
\newfloat{listing}{tb}{lst}{}
\floatname{listing}{Listing}

\title{Low-Light Video Enhancement with Synthetic Event Guidance}
\author{Lin Liu\textsuperscript{1} \hspace{1mm}
Junfeng An\textsuperscript{2} \hspace{1mm}
Jianzhuang Liu\textsuperscript{3}\hspace{1mm}
Shanxin Yuan\textsuperscript{3}\thanks{Corresponding author}\hspace{1mm}
Xiangyu Chen\textsuperscript{5} \hspace{1mm} \\
Wengang Zhou\textsuperscript{1} \hspace{1mm}
Houqiang Li\textsuperscript{1} \hspace{1mm}
Yanfeng Wang\textsuperscript{6} \hspace{1mm}
Qi Tian\textsuperscript{4} \\
 \footnotesize{$^1$University of Science and Technology of China} \quad
 \footnotesize{$^2$ Harbin Institute of Technology} \\
 \footnotesize{$^3$Huawei Noah's Ark Lab} \quad
 \footnotesize{$^4$Huawei Cloud BU} \quad
 \footnotesize{$^5$University of Macau} \quad
 \footnotesize{$^6$Shanghai AI Laboratory}
}

\usepackage{bibentry}

\begin{document}

\maketitle

\begin{abstract}
Low-light video enhancement (LLVE) is an important yet challenging task with many applications such as photographing and autonomous driving.
Unlike single image low-light enhancement, most LLVE methods utilize temporal information from adjacent frames to restore the color and remove the noise of the target frame.
However, these algorithms, based on the framework of multi-frame alignment and enhancement, may produce multi-frame fusion artifacts when encountering extreme low light or fast motion.
In this paper, inspired by the low latency and high dynamic range of events, we use synthetic events from multiple frames to guide the enhancement and restoration of low-light videos.
Our method contains three stages: 1) event synthesis and enhancement, 2) event and image fusion, and 3) low-light enhancement. In this framework, we design two novel modules (event-image fusion transform and event-guided dual branch) for the second and third stages, respectively. Extensive experiments show that our method outperforms existing low-light video or single image enhancement approaches on both synthetic and real LLVE datasets.
\end{abstract}

\begin{figure}[!t]
		\centering
		\includegraphics[ width=0.48\textwidth]{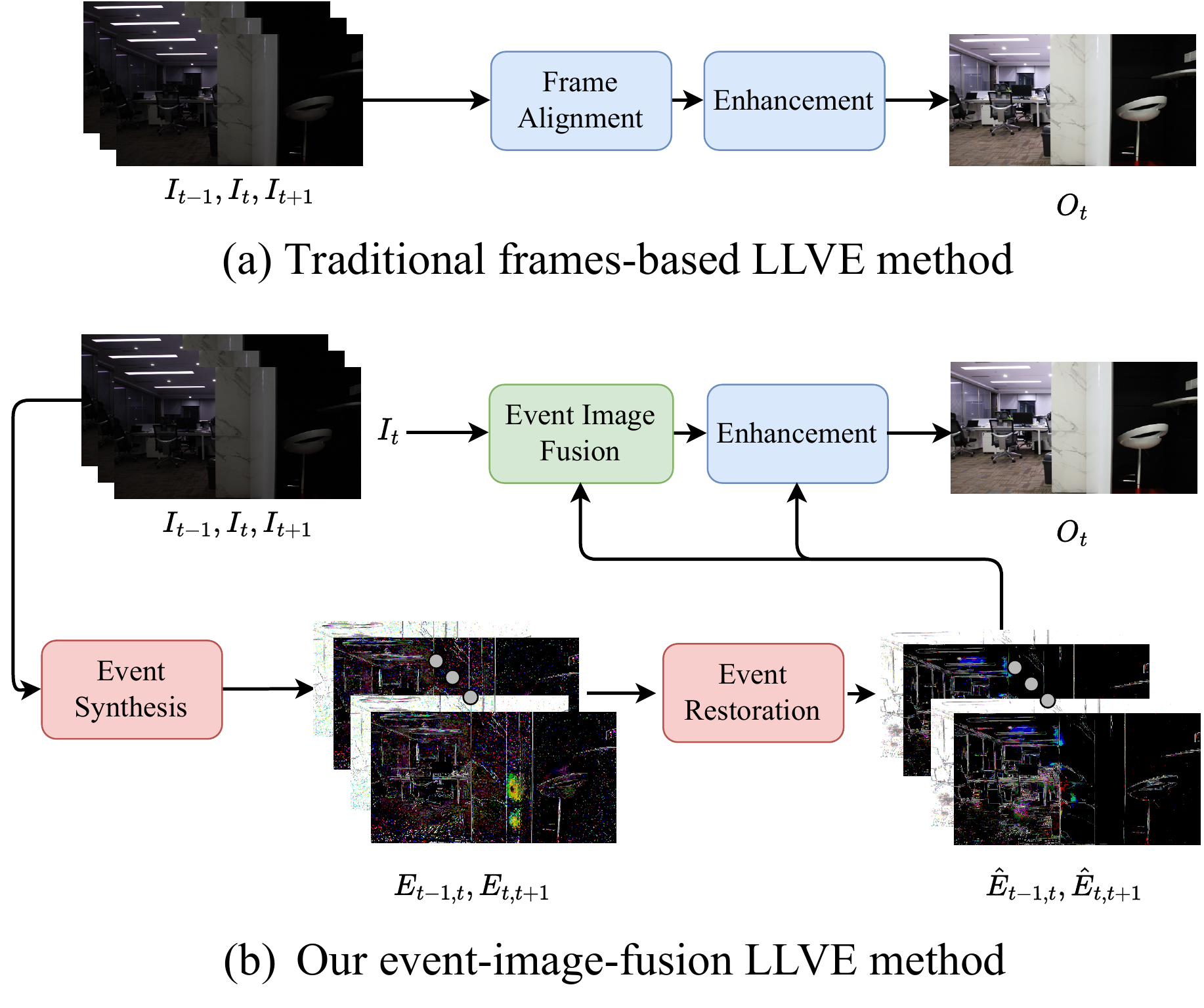}
		\includegraphics[ width=0.48\textwidth]{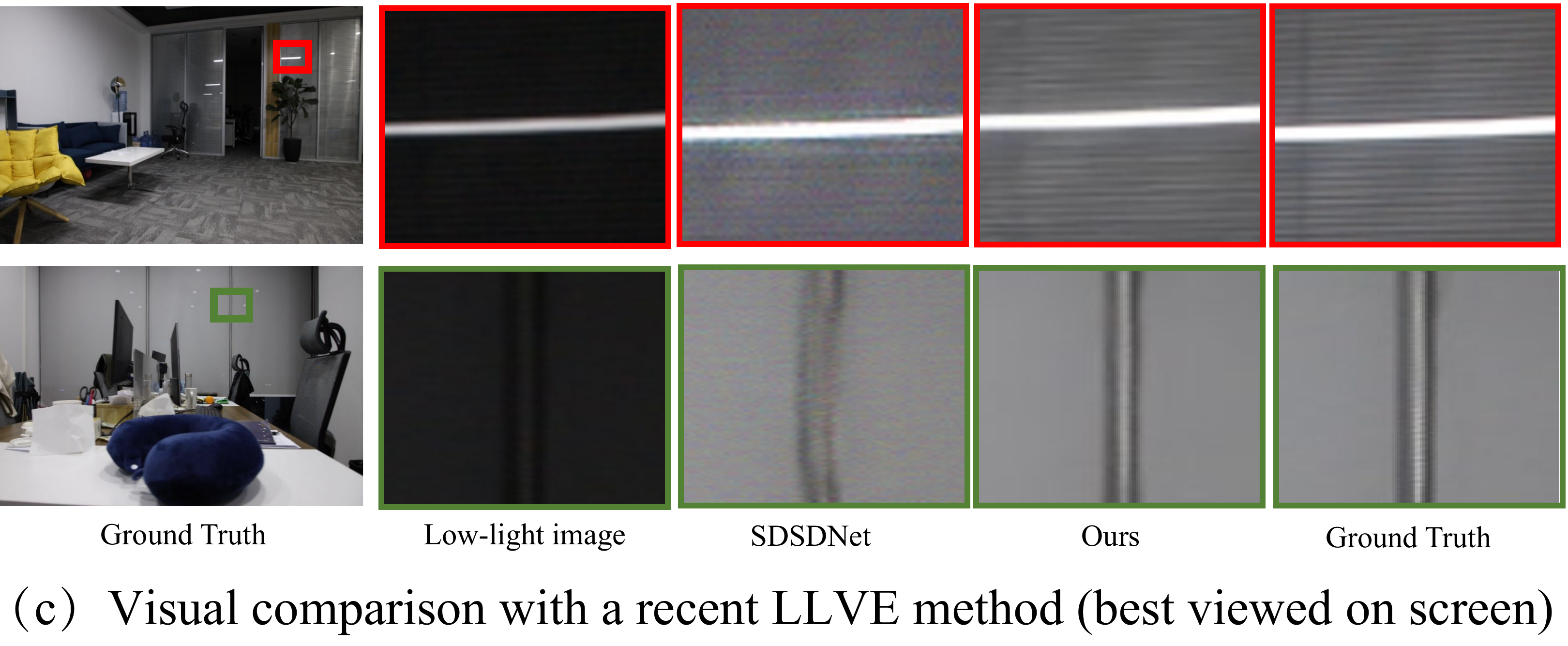}
		\caption{Comparison between (a) Traditional frames-based LLVE method and (b) Our event-image fusion LLVE method. (c) Visual comparison between a recent LLVE method SDSDNet~\cite{RuixingWang2021SeeingDS} and ours. Our method can better remove noise, maintain details and avoid misaligned artifacts of multi-frame fusion.} 
		\label{fig:index}
\end{figure}

\section{Introduction}
The image quality in low-light or under-exposure conditions is often unsatisfactory, so image/video enhancement in low light has been an active research topic in computer vision.
However, it is challenging due to strong noise, detail loss, non-uniform exposure, \textit{etc}.
These problems become even more serious in videos taken from dynamic scenes. In this paper, in contrast to low-light, we loosely call bright-light and day-light images/videos/events \textit{normal-light} images/videos/events.

Most of fully-supervised deep-learning based low-light video enhancement (LLVE) methods~\cite{FeifanLv2018MBLLENLI,HaiyangJiang2019LearningTS,RuixingWang2021SeeingDS} or video reconstruction methods~\cite{TianfanXue2019VideoEW,XintaoWang2019EDVRVR,TakashiIsobe2020VideoSW,PengDai2022VideoDW} are based on the multi-frame alignment-and-enhancement framework.
This pipeline firstly utilizes some techniques, \textit{e.g.}, 3D convolution~\cite{FeifanLv2018MBLLENLI,HaiyangJiang2019LearningTS}, deformable convolution~\cite{TakashiIsobe2020VideoSW,PengDai2022VideoDW}, or flow-based alignment~\cite{TianfanXue2019VideoEW}, to align the temporal information from adjacent frames to the reference frame, and then uses an enhancement network for noise removal and illumination correction (see Fig.~\ref{fig:index}(a)).
However, when facing extreme low light or significant motion in videos, these algorithms may produce multi-frame fusion artifacts in the predicted images (see the results of SDSDNet in Fig.~\ref{fig:index}(c)).

Existing methods face some potential difficulties. First, sensor noise is not negligible in low signal-to-noise low-light scenarios.
This noise hinders the network from learning the alignment of temporal features.
Second, the interference between strong noise and image details causes the enhancement network to remove some image details inevitably.

In this paper, inspired by the low latency and high dynamic range of events, we use synthetic
events to guide the enhancement and restoration of low-light videos.
In general, events are captured by event cameras (e.g., DAVIS240C~\cite{EliasMueggler2017TheED}), which contain sparse and asynchronous intensity changes of the scene, instead of the color and intensity as in normal images.
Recently, Gehrig \textit{et al.}~\cite{Gehrig_2020_CVPR} present a method that can convert videos recorded with conventional cameras into synthetic realistic
events.
The pioneering work applying event information to low-light enhancement is SIDE ~\cite{SongZhang2020LearningTS}. 
However, their work studies the transformation from real events to single images, and due to the difficulty of collecting real event-image pairs, it is an unpaired learning method.
Differently, we focus on the fusion between synthetic enhanced events and video frames, where the synthetic events and frames are paired.

\begin{figure}[!t]
		\centering
		\includegraphics[
		 width=0.48\textwidth]{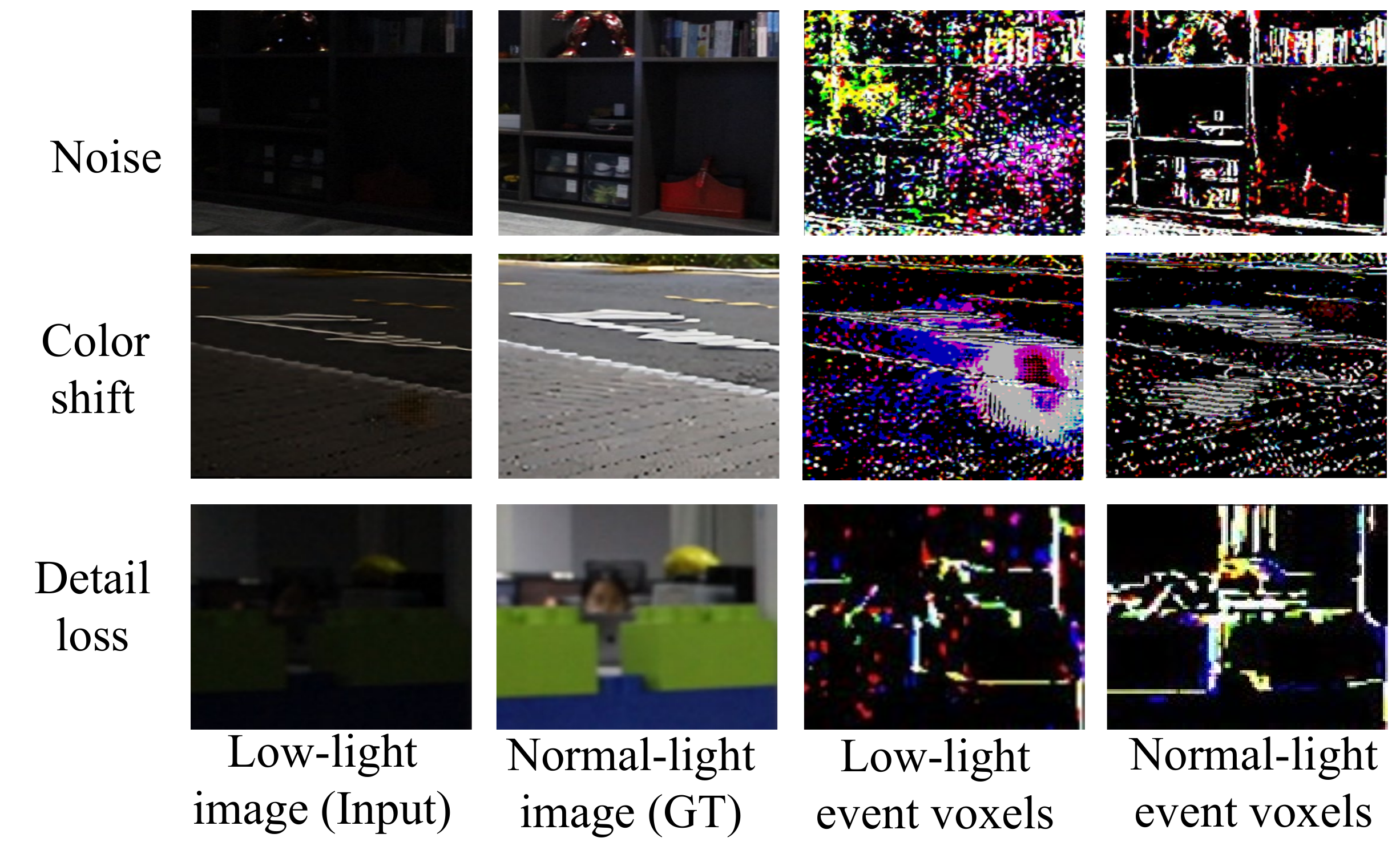}
		\caption{Challenges of low-light image/video enhancement. When comparing the events of normal-light images and low-light images, we find that low-light enhancement faces challenges of strong noise, color shift, and detail loss. These
       problems are hard to notice in normal-light images, but are apparent in the event voxels.} 
		\label{fig:challengs}
\end{figure}

Unlike the conventional two-stage video reconstruction/enhancement pipeline, we propose a three-stage LLVE pipeline with the guidance of synthetic events: 1) event synthesis and restoration; 2) event-image fusion; 3) low-light enhancement (see Fig.~\ref{fig:index}(b)).
The first stage is event-to-event, aiming to obtain normal-light events from low-light events.
Due to the sparsity of events and the interpolation of fast motion by an event synthesis algorithm, our network can restore normal-light events well and solve the problems of noise, color shift, and detail loss based on the enhanced events (see Fig.~\ref{fig:challengs}).

Different from most event-image fusion methods~\cite{StefanoPini2018LearnTS,StefanoPini2019VideoSF,JinHan2020NeuromorphicCG,GenadyPaikin2021EFINetVF,he2022timereplayer}, which combine event and image features by simply concatenating them, we design a novel Event and Image Fusion Transform (EIFT) module for better fusing the sparse event features and image features in the second stage of our pipeline.
And in the last stage, we design an Event-Guided Dual Branch (EGDB) module
for low-light enhancement.
Using the mask generated by the enhanced events, the image can be divided into areas with little change and areas with sharp change during a period of time.
Moreover, the proposed global branch and local branch in EGDB can deal with these two kinds of areas, respectively.
A transformer network is used in the global branch to capture global illumination information.

In summary, we make the following contributions:
\begin{itemize}
  \item A novel three-stage pipeline with the guidance of synthetic events for low-light video enhancement is proposed. It consists of event synthesis and restoration, event image fusion, and low-light enhancement. 
 
  \item We design a novel Event and Image Fusion Transform (EIFT) module for event-image fusion and an Event-Guided Dual Branch (EGDB) module for low-light enhancement. 
  
  \item Extensive experiments on both synthetic and real LLVE datasets show that our method outperforms both the state-of-the-art (SOTA) low-light video and single image enhancement approaches.
  
\end{itemize}

\section{Related Work}

This section discusses the most related methods, including low-light video enhancement, event-image fusion, and low-level vision Transformers.

\noindent
\textbf{Low-Light Video Enhancement (LLVE).}
This line of work focuses on how to utilize temporal information from adjacent frames. They can be divided into paired learning and unpaired learning.
In the former, most methods use the alignment-and-enhance framework for LLVE.
Lv \textit{et al.}~\cite{FeifanLv2018MBLLENLI} and Chen \textit{et al.}~\cite{ChenChen2019SeeingMI} use 3D convolution and their networks can combine temporal information from image sequences.
Wang \textit{et al.} use a mechatronic system to collect a dataset named SDSD, where the normal-light and low-light image pairs are well aligned~\cite{RuixingWang2021SeeingDS}. 
They also use deformable convolutions in the multi-frame alignment stage.
Some approaches try to solve the LLVE problem by combining an image-based model with temporal consistency loss~\cite{ChenChen2019SeeingMI,SongZhang2020LearningTS}.
However, most of the temporal consistency losses in LLVE~\cite{FanZhang2022LearningTC} or other video restoration tasks~\cite{MingHsuanYang2018LearningBV,HengyuanZhao2021TemporallyCV} are optical-flow based, which may be limited by the error in optical flow estimation.
As for unpaired learning methods, to solve the lack of data problem, Triantafyllidou \textit{et al.}~\cite{triantafyllidou2020low} convert normal-light videos to low-light videos using two CycleGAN~\cite{JunYanZhu2017UnpairedIT} networks.
In this framework, normal-light videos are first converted into long exposure videos and then into low-light videos. 

\noindent
\textbf{Event-Based Image/Video Reconstruction.}
The event camera can handle high-speed motion and high-dynamic scenes because of its dynamic vision sensor.
Therefore, events can be applied to image or video reconstruction tasks including video deblurring~\cite{jiang2020learning}, super-resolution~\cite{han2021evintsr,jing2021turning}, joint filtering~\cite{wang2020joint}, and tone mapping~\cite{simon2016event,JinHan2020NeuromorphicCG}.
Our work is the first to explore the application of synthetic events in the low-light video enhancement task.

\noindent
\textbf{Event-Image Fusion.}
Event features and image features are of different modals, with different characteristics. 
Event features reflect motion changes, so most values in the scene at a certain moment are zero.
Image features in low light contain both strong noise and scene structure.
Most works simply concatenate or multiply event features and image features~\cite{StefanoPini2018LearnTS,StefanoPini2019VideoSF,JinHan2020NeuromorphicCG,GenadyPaikin2021EFINetVF,he2022timereplayer,Tomy2022fusingevent}.
Han \textit{et al.}~\cite{JinHan2020NeuromorphicCG} fuse the low-dynamic-range image features and the event features for high-dynamic-range image reconstruction.
He \textit{et al}.~\cite{he2022timereplayer} concatenate the optical flow, event, and image together for video interpolation.
Tomy \textit{et al.}~\cite{Tomy2022fusingevent} multiply the multi-scale image and event features for robust object detection.
Very recently, Cho \textit{et al.}~\cite{cho2022Eif} design an event image fusion module for depth estimation.
The differences between these methods and our work are: 1) we deal with the LLVE task; 2) they mainly use image features to densify sparse event features, while in our work, event and image features enhance each other for restoration.

\noindent
\textbf{Low-Level Vision Transformers.} 
In recent years, Transformers have been applied in many low-level visual tasks, and they can be divided into multi-task and task-specific Transformers.
For multi-task Transformers, Chen \textit{et al.}~\cite{chen2020pre} and Li \textit{et al.}~\cite{li2021efficient} propose IPT and EDT, respectively, which are pre-trained on several low-level vision tasks.
Wang \textit{et al.}~\cite{ZhendongWang2021UformerAG} and Zamir~\cite{Zamir2021Restormer} design U-Net-like Transformer architectures for denoising, deblurring, deraining, \textit{etc}.
Task-specific Transformers also obtain state-of-the-art performances in many applications including super-resolution~\cite{FuzhiYang2020LearningTT,chen2022activating}, deraining~\cite{Jie2022idt,Liang_2022_CVPRw}, and inpainting~\cite{yan2020sttn}.
In our paper, we use Transformer in the enhancement stage to deal with the areas in the videos without fast motion.

\section{Method}
In this section, we first describe our overall pipeline and then present the three stages of our method, including event synthesis and restoration, event-image fusion, and low-light enhancement.

\begin{figure}[!t]
		\centering
		\includegraphics[
		 width=0.47\textwidth]{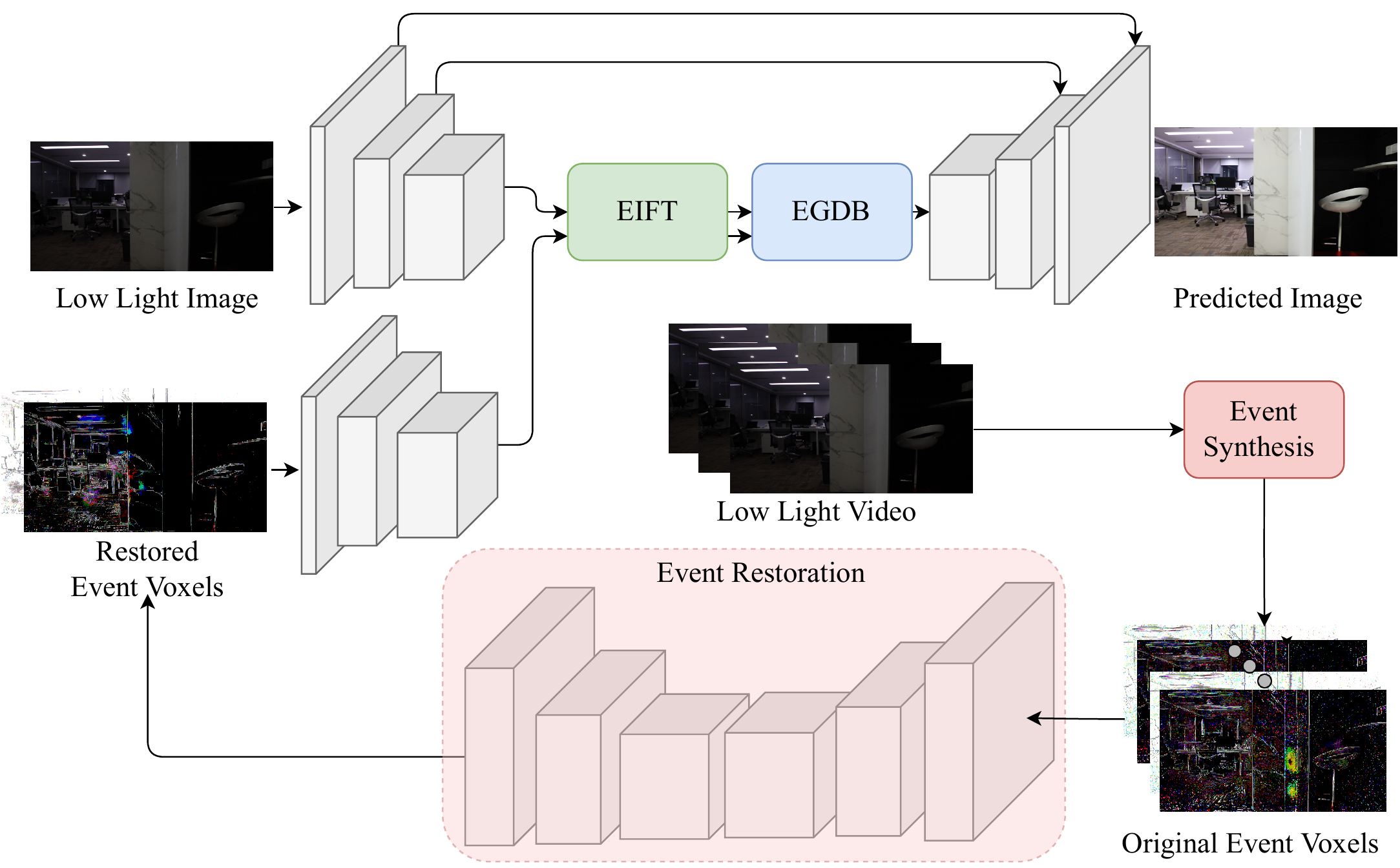}
		\caption{Architecture of our overall network. The original event voxels pass the event restoration network and generate the restored event voxels. And the low-light image and the restored event voxels are combined to obtain the final enhanced normal-light results.} 
		\label{fig:overallnet}
\end{figure}

As shown in Fig.~\ref{fig:overallnet}, firstly, the original event voxels are obtained by synthesizing light sequences (Sec. 3.1). And the restored event voxels are generated from a U-Net (Sec. 3.1).
The image and the restored event voxels are encoded to deep features, fused by the EIFT module (Sec. 3.2) and enhanced by the EGDB module (Sec. 3.3). 
Finally, the decoder outputs the enhanced images. 

\subsection{Events Synthesis and Restoration}
\textbf{Event Synthesis.} To synthesize original event voxels, we have the following three steps: frame upsampling, event generation, and event voxel generation.
In the \textbf{first} step, we use an off-the-shelf up-sampling algorithm~\cite{xiang2020zooming} to get $N'$ up-sampled frames (with $N'$ adaptively chosen) from $N$ frames ($N' > N$).
In the \textbf{second} step, two adjacent frames (from $N'$ frames) are subtracted to obtain $d_{x,y,t}$, and whether an event $e_{i}$ is generated is determined by the difference.
If $d_{x_{i}, y_{i}, t_{i}}$ exceeds a threshold, we generate an event $e_{i} = \left(x_{i}, y_{i}, t_{i}, p_{i}\right)$, where $(x_{i}, y_{i})$ and $t_{i}$ are the location and time of the event and $p_{i} = \pm 1$ $(p_{i}  d_{x_{i}, y_{i}, t_{i}} > 0)$.
We use one threshold for low-light events (set to 2) and another for normal-light events (set to 5).
This is because setting a higher threshold when generating low-light events will lose some useful events.
In the \textbf{third} step, in order to make the events processed by neural networks better, it is necessary to convert the discrete event values $\{0,+1,-1\}$ into floating-point event voxels.
Following ~\cite{zhu2019unsupervised,weng2021event}, we generate the voxels $E \in \mathbb{R}^{B \times H \times W}$ as:
\begin{small} 
\begin{equation}
E_{k}\! =\! \sum^{n}_{i=0} p_{i} \max \left( \! 0,1 \! - \!\left|k \! - \! \frac{t_{i}\! - \! t_{0}}{t_{n}\! -\! t_{0}}(B \! - \! 1)\right|\right), \ k\in \{1,...,B\},
\label{eq:evvoxel}
\end{equation}
\end{small}%
where $t_{0}$ and $t_{n}$ denote the start and end moments of the events, respectively. $B$ equals to $2N \times 3$, where $N$ is the input frame number, 2 corresponds to the positive and negative event voxels, and 3 corresponds to the r-g-b color channels.
This equation shows that we use temporal bilinear interpolation to obtain $B$ voxels of the event temporal information.

\textbf{Event Restoration.} In the event restoration, we aim to generate restored events from the original events using an CNN.
To the best of our knowledge, there is no available neural network to transfer low-light events to normal-light event voxels.
We design a network (see the supplementary materials) that is able to predict an event probability map $ P \in \mathbb{R}^{B \times H \times W}$ and an event voxel map $V\in \mathbb{R}^{B \times H \times W} $ simultaneously.
Finally, the restored event voxels are calculated as:
\begin{equation}
E^{r}=M(P) V ,
\end{equation}
where the element $M_{i}(P_{i})$ of $M(P)$ is 1 when $P_{i} \ge 0.5$ and 0 otherwise.
The loss functions that constrain $M$ and $V$ will be described in Sec. 3.4.

\begin{figure*}[!t]
		\centering
		
		\includegraphics[ width=0.88\textwidth]{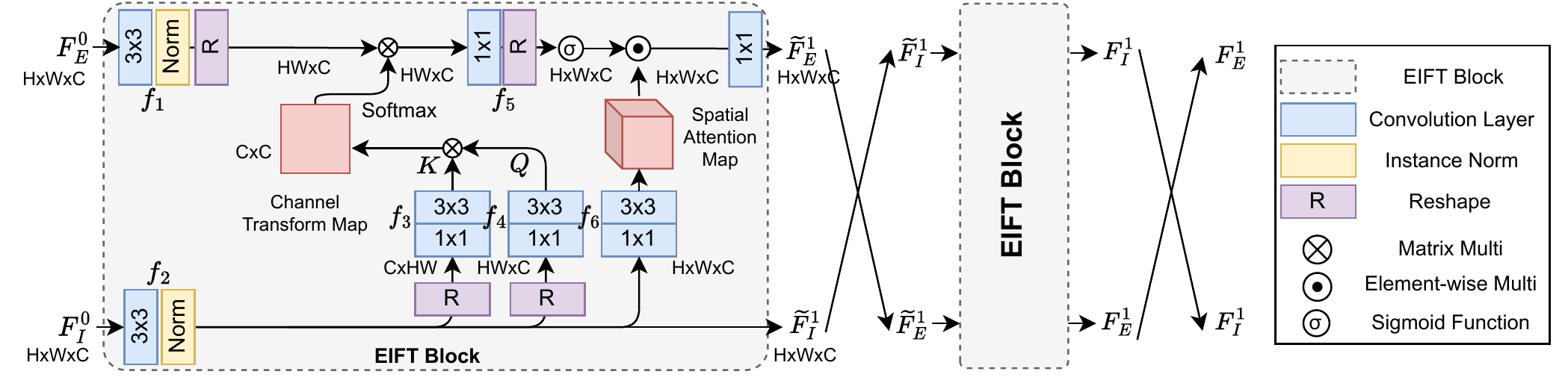}
		\caption{Architecture of the EIFT module. All the ReLU layers are omitted for clarity.} 
		\label{fig:fusionblock}
\end{figure*}

\subsection{Event-Image Fusion}
Different from most event-image fusion methods~\cite{StefanoPini2018LearnTS,StefanoPini2019VideoSF,JinHan2020NeuromorphicCG,GenadyPaikin2021EFINetVF,he2022timereplayer}, which combine event and image features by concatenating them simply, we design a novel Event and Image Fusion Transform (EIFT) module for better fusing the sparse event features and dense image features.
Our EIFT is based on the mutual attention mechanism, where the event features and the image features modulate each other.
Specifically, the event feature modulates the image feature by hinting where fast motion happens and distinguishing between strong noise and image details; the image feature modulates the event feature by introducing the color or semantic information.
In our EIFT, such modulation is achieved by generating the channel transform map and the spatial attention map from the modulated features. The maps have values between 0 and 1 indicating the importance of every element.

Fig.~\ref{fig:fusionblock} is the designed network structure of EIFT. Each EIFT module contains two EIFT blocks.
In the first block, $F^{0}_{E}$ and $F^{0}_{I}$ serve as the main feature and the modulation feature, respectively.
Inspired by~\cite{Zamir2021Restormer}, to reduce computational complexity, our modulation is divided into cross-channel transform (CCT) and element-wise product (EWP).
In CCT, the modulation feature passes two parallel convolution blocks and generates $
{Q} \in \mathbb{R}^{H W \times C}$ and $ {K} \in \mathbb{R}^{C \times H W}$. 
The dot product of $K$ and $Q$ is performed, and the Softmax function is applied to generate a $C \times C$ channel transform map. Then it is multiplied by the main feature. The whole CCT can be formulated as: 
\begin{equation}
{X} = f_{1}(F_{E}) \cdot \operatorname{Softmax}\{f_{3}(f_{2}(F_{I})) \cdot f_{4}(f_{2}(F_{I}))\}.
\label{eq:s1}
\end{equation}

In EWP, the modulation feature first passes through some convolution blocks to generate a spatial attention map. And the element-wise multiplication product between the spatial attention map and the main feature is carried out, which can be formulated as:
\begin{equation}
F=\sigma\left(f_{5}(X)\right) \odot f_{6}(f_{2}(F_{I})).
\label{eq:s2}
\end{equation}
$f_{i}$ in Eqn.~\ref{eq:s1} and Eqn.~\ref{eq:s2} are the convolution blocks indicated in Fig.~\ref{fig:fusionblock}. $\sigma$ and $\odot$ denote the Sigmoid function and the element-wise production, respectively.

In the second EIFT block, the inputs (outputs of the first EIFT block) are swapped, where $\tilde{F}^{n}_{E}$ is the modulation feature and $\tilde{F}^{n}_{I}$ is the main feature. The final outputs of the EIFT module are $F^{n}_{E}$ and $F^{n}_{I}$ for the $n$-th EIFT module. In this paper, $n$ is set to $2$.

\subsection{Low-Light Enhancement}

In low-light image/video enhancement, illumination estimation/enhancement is important.
The previous LLVE methods often process the whole image using convolution networks.
However, the difficulties for illumination enhancement in areas with little motion and in areas with fast motion are quite different.
Especially when there is fast motion, where the illumination will change, the estimated illumination is often inaccurate.
We make two improvements: 1) we use two branches to deal with these two kinds of areas (with fast motion, and without fast motion) respectively for illumination enhancement; 2) we use a Transformer to enhance the brightness.

\textbf{Mask Generation.} Given the restored event voxels, $E^{r} \in \mathbb{R}^{(2N \times 3) \times H\times W}$, generated from the first stage (Sec. 3.1), we use the positive voxels $E^{r,+} \in \mathbb{R}^{(N \times 3) \times H\times W}$ for generating the mask $M'$, which is computed by:
\begin{equation}
\begin{aligned}
M_{c} = \operatorname{max}\{ E^{r,+}_{c,1},E^{r,+}_{c,2},...,E^{r,+}_{c,N}\}, \ c \in \{r,g, b\},
\end{aligned}
\end{equation}
\begin{equation}
M' = \operatorname{max}\{M_{r},M_{g},M_{b}\} \in \mathbb{R}^{H \times W},
\end{equation}
where $M'$ is then resized to the same resolution as the input of EGDB.
Finally, $M'$ is changed to a binary 0-1 mask with a threshold 0.9.

\begin{figure}[!t]
		\centering
		
		\includegraphics[ width=0.47\textwidth]{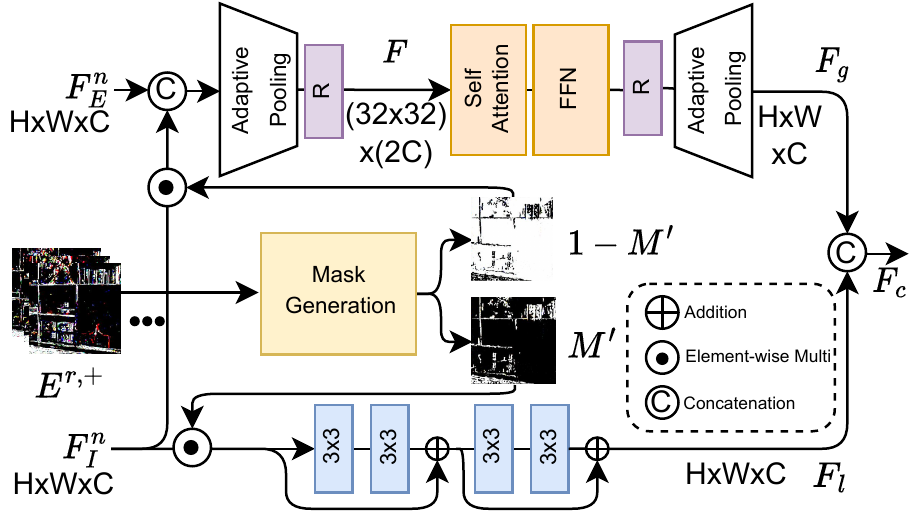}
		\caption{Architecture of the EGDB module with the global branch (upper part) and the local branch (bottom part). All the ReLU layers are omitted for clarity.}  
		\label{fig:enhancementnet}
\end{figure}

\textbf{Network Details.} In Fig.~\ref{fig:enhancementnet}, EGDB consists of a global branch and a local branch.
In the global branch, we first concatenate the $F^{n}_{E}$ and the masked image feature $(1-M')F^{n}_{I}$ from the event-image fusion stage and perform the adaptive average pooling to get the feature $F \in \mathbb{R}^{32 \times 32 \times 2C}$. 
Because of the color uniformity and texture sparsity of these kinds of areas, down-sampling can reduce network parameters without loss of performance.
The feature $F$ is then reshaped, splited into $m$ feature patches $F_{i} \in \mathbb{R}^{ p ^{2} \times 2C}, i=\{1, \ldots, m\}$, where $p$ is the patch size and $m=\frac{H}{p} \times \frac{W}{p}$, and passes a self-attention layer and a feed-forward layer of the conventional Transformer~\footnote{See the supplementary materials for more network details.}.
The global branch outputs $F_{g}$ with the same size of $F^{n}_{E}$.
The local branch with two residual blocks receives $M'F^{n}_{I}$ and outputs the feature $F_{l}$.
Finally, the EGDB module outputs $F_{c}$, which is the concatenation of $F_{g}$ and $F_{l}$.

\subsection{Training and Loss Functions}
The training contains two stages.
In the first stage, with the low-light event voxels and the normal-light event voxels, we train the event restoration network using
\begin{equation}
L_{s1}=L_{m}+\lambda_{1} L_{v},
\end{equation}
where the first term is the binary cross-entropy loss between the predicted event probability map $P$ and the binary normal-light voxel mask $M^{G}$ (generated by the normal-light event voxel ground truth $G$ with a threshold 0.1):
\begin{equation}
L_{m}\!=\!-\frac{1}{H W}\! \sum_{i=1}^{H W} M^{G}_{i} \! \log \left(P_{i}\right)+\left(1 \! -\! M^{G}_{i}\right) \! \log \left(1\! -\! P_{i}\right),
\end{equation}
where $P_{i} \in P$ and $M^{G}_{i} \in M^{G}$. The second term $L_{v}$ is the $L_{1}$ loss between $E^{r}$ and $G$.

After the first stage, the parameters of the event restoration network are fixed, and the other networks are trained. The loss function is defined as:
\begin{equation}
L_{s2}=L_{1}+\lambda_{2} L_{vgg},
\end{equation}
where $L_{1}$ and $L_{vgg}$ are respectively the $L1$ loss and the perceptual loss~\cite{johnson2016perceptual} between the predicted normal-light image and its ground truth.

\label{sec:basicnetwork}

\section{Experiments and Analysis}
\label{sec:exp}
In this section, we conduct an ablation study and compare with state-of-the-art methods.

\subsection{Datasets and Implementation Details}
\textbf{Datasets.} For a real low-light video dataset, we adopt SDSD~\cite{RuixingWang2021SeeingDS} which contains 37,500 low- and normal-light image pairs with dynamic scenes. 
SDSD has an indoor subset and an outdoor subset.
For a fair comparison, we use the same training/test split as in~\cite{RuixingWang2021SeeingDS}.
We do not consider the real LLVE datasets which are not released~\cite{HaiyangJiang2019LearningTS,WangWei2019EnhancingLL} or only contain static scenes~\cite{ChenChen2019SeeingMI}.
In order to have experiments on more scenes and data, following the work~\cite{triantafyllidou2020low}, we also perform experiments on Vimeo90k~\cite{TianfanXue2019VideoEW}. 
We select video clips whose average brightness is greater than 0.3 (the highest brightness is 1) as normal-light videos, and then use the method in ~\cite{lv2021attention} to synthesize low-light video sequences~\footnote{More details are shown in the supplementary materials.}.
We finally get 9,477 training and 1,063 testing sequences, each with 7 frames, from Vimeo90k.

\textbf{Implementation Details.} We implement our method in the PyTorch~\cite{AdamPaszke2019PyTorchAI} framework, and train and test it on two 3090Ti GPUs. 
The network parameters are randomly
initialized with the Gaussian distribution.
In the training stage, the patch size and batch size are 256 and 4, respectively.
We adopt the Adam~\cite{DiederikPKingma2014AdamAM} optimizer with momentum set to 0.9.
The input number of frames $N$ is set to 5.

\begin{table}[t]
\centering\small
  \vspace{-2mm}
\resizebox{8.6cm}{!}{
\begin{tabular}{clcccc}
\toprule
\multicolumn{1}{l}{}                                                           &         & PSNR$\uparrow$  & SSIM$\uparrow$ & LPIPS$\downarrow$ & Network Size \\ \midrule
\multirow{8}{*}{\begin{tabular}[c]{c}Image-based\\ method\end{tabular}}  & DeepUPE & 21.82 & 0.68 & --      &     0.59M   \\
                                                                               & ZeroDCE & 20.06 & 0.61 &  --     &   0.08M     \\
                                                                               & DeepLPF & 22.48 & 0.66 &  --     &   1.77M    \\
                                                                               & DRBN    & 22.31 & 0.65 &   --    &  1.12M \\
                                                                               & Uformer* & 23.46 & 0.72 &  0.202     &  20.4M        \\
                                                                               & STAR*    &  23.39     &   0.70   & 0.283  &  0.03M            \\
                                                                               & SCI*      &  19.67     &   0.69  &  0.298    &  0.01M            \\
                            & LLFlow*      &  24.90     &   0.78  &   0.182   &  5.43M         \\                                              
                                                                               \midrule
\multirow{5}{*}{\begin{tabular}[c]{c}Video-based \\ method\end{tabular}} & MBLLVEN & 21.79 & 0.65 &  0.190     &   1.02M           \\
                                                                               & SMID    & 24.09 & 0.69 &   0.213    &   6.22M     \\
                                                                               & SMOID   & 23.45 & 0.69 & 0.187   &  3.64M            \\
                                                                               & SDSDNet & 24.92 & 0.73 & 0.138  &          4.43M    \\
                                                                               & SGZSL*   &  23.89 &  0.70    & 0.308   &   28.1M     \\
                         \midrule
\multicolumn{1}{l}{}                                                           & Ours    & \bf{25.81} & \bf{0.80} &  \bf{0.126}  & 3.51M  \\
\bottomrule
\end{tabular}}
  \caption{Quantitative frame-based low-light enhancement comparison on SDSD~\cite{RuixingWang2021SeeingDS}. The best results are in \textbf{bold}. * denotes methods implemented by us using their official models and other results are cited from~\cite{RuixingWang2021SeeingDS}.}
\end{table}

\begin{table}[t]
  \centering\small
  \resizebox{7.9cm}{!} {
  \begin{tabular}{c|cccc}
    \toprule
      Model&PSNR$\uparrow$ & SSIM $\uparrow$ & LPIPS $\downarrow$ & Network size\\
    \midrule
      ZeroDCE&24.56 & 0.7533& 0.202 & 0.08M\\
      STAR& 24.72 & 0.7595 & 0.195 & 0.03M \\
      Uformer& 30.29 & 0.9203 &  0.058 & 20.4M\\
      MBLLVEN&27.06 & 0.8706 &0.083 & 1.02M \\
      SMOID& 29.74 & 0.9239 & 0.059& 3.64M \\
      SDSDNet&29.06& 0.9233&  0.064& 4.43M\\
      Ours& \textbf{30.53} & \textbf{0.9254} & \textbf{0.038}   & 3.51M \\
    \bottomrule
  \end{tabular}}
  \label{tab:vimeo90k}
\caption{Quantitative frame-based low-light enhancement comparison on the Vimeo90K dataset.}
\end{table}

\begin{table}[t]
  \centering\small

  \resizebox{5.2cm}{!} {
  \begin{tabular}{c|ccc}
    \toprule
      Model&FID$\downarrow$&Warping Error$\downarrow$\\
      \midrule
      MBLLVEN&0.533& 2.98 \\
      SMID & 0.552 & 2.85 \\
      SMOID& 0.521& \underline{2.62}\\ 
      SDSDNet&\textbf{0.368}&4.36\\ 
      SGZSL&0.722&6.13 \\ 
      Ours& \underline{0.370}& \textbf{2.61}\\ 
    \bottomrule
    
  \end{tabular}}
    \caption{Quantitative video-based low-light enhancement
comparison on the SDSD dataset. The best results are in \textbf{bold} and the second best are \underline{underlined}.}
  \label{tab:videosdsd}
\end{table}

 \begin{figure*}[!t]
		\centering
		
		\includegraphics[ width=0.91\textwidth]{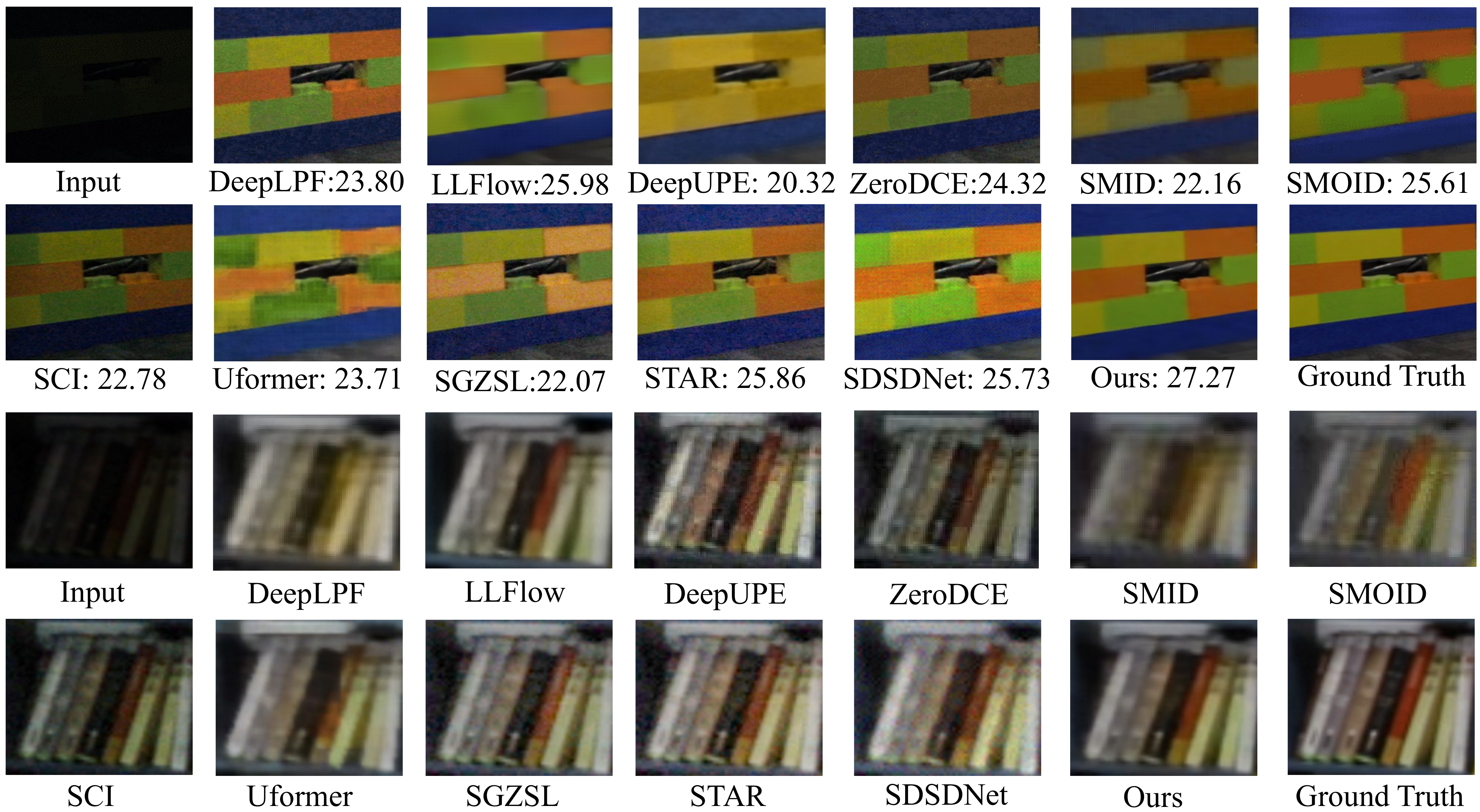}
		\caption{Visual comparison between other SOTA methods and ours on two patches of one example from SDSD. The numbers are PSNR values on the whole image.}  
		\label{fig:resultsdsd}
\end{figure*}

\subsection{Comparison with State-of-the-Arts.}\textbf{State-of-the-Art Methods.}
We compare our work with seven low-light single image enhancement methods (DeepUPE~\cite{RuixingWang2019UnderexposedPE}, ZeroDCE~\cite{ChunleGuo2020ZeroReferenceDC}, DeepLPF~\cite{SeanMoran2020DeepLPFDL}, DRBN~\cite{WenhanYang2020FromFT}, STAR~\cite{ZhaoyangZhang2021STARAS}, SCI~\cite{LongMa2022TowardFF} and LLFlow~\cite{YufeiWang2022LowLightIE}, one SOTA transformer-based general image restoration method (Uformer~\cite{ZhendongWang2021UformerAG}), and five low-light video enhancement methods (MBLLVEN~\cite{FeifanLv2018MBLLENLI}, SMID~\cite{ChenChen2019SeeingMI}, SMOID~\cite{HaiyangJiang2019LearningTS}, SDSDNet~\cite{RuixingWang2021SeeingDS} and SGZSL~\cite{ShenZheng2022SemanticGuidedZL}).
In these methods, SCI and Zero-DCE are un-supervised, and the models marked by $*$ in Table 1 and all the models in Table 2 are implemented using their official models on the corresponding datasets.

\textbf{Quantitative Results.} 
For frame-based comparison, we use PSNR and SSIM~\cite{ZhouWang2004ImageQA}, and Learned Perceptual Image
Patch Similar (LPIPS)~\cite{zhang2018perceptual} to compare the restored images. 
For video-based comparison, we adopt FID and Warping Error for video-based methods. These two metrics can well reflect the quality and stability of the predicted videos.
As shown in Table 1 and Table 2, our method obtains the best performances on both SDSD and Vimeo90K for frame-based comparisons.
For video-based comparison, our method gets the 2nd on FID (only 0.002 lower than the best method) and the 1st on Warping Error without using any stability loss like SMID.

\textbf{Qualitative Results.} In Fig.~\ref{fig:resultsdsd}, we show visual comparisions on SDSD. Most single-image enhancement methods either produce some noise (DeepLPF, ZeroDCE, Uformer and STAR) or have color shift (DeepUPE and SCI).  
As for the video-based methods, SMID does not restore the color well, SMOID blurs the details. The deviation of noise map estimation by SDSDNet results in a certain degree of noise in the prediction.
With the help of events, our method restores the color and removes the noise simultaneously.
We show other visual results on Vimeo90K in the supplementary materials, where ours also outperforms the others.

\begin{table}[t]
  \centering\small
  
  \vspace{-2mm}
  \resizebox{6.9cm}{!} {
  \begin{tabular}{c|ccc}
    \toprule
      Model&PSNR $\uparrow$&SSIM$\uparrow$&LPIPS$\downarrow$\\
      \midrule
      EIFT $\rightarrow$ UNet & 25.54 & 0.79& 0.135\\
      W/o CCT&25.46 & 0.78 & 0.141\\
      W/o EWP &25.52 &0.79 &0.138\\
      \midrule
      W/o Event Guidance &25.49 & 0.78 & 0.143\\
      W/o Global Branch &25.34 & 0.77 & 0.167\\
      W/o Local Branch &25.42 & 0.78 & 0.162 \\
      \midrule
      PCD+EGDB (W/o EG) & 24.27 & 0.73 & 0.185\\
       \midrule
      Full Model & \bf{25.81} & \bf{0.80} & \bf{0.126}\\
      
    \bottomrule
    
  \end{tabular}}
  \vspace{-2mm}
  \caption{Ablation study on SDSD. EG: event guidance.}
  \label{tab:abstudy}
\end{table}

\subsection{Ablation Study}
In this section, we do an ablation study based on the SDSD dataset and show the results in Table~\ref{tab:abstudy}.

\subsubsection{Network Architecture.}
First, we analyze the effectiveness of the network architecture.

\textbf{EIFT}. We construct three models: 1) EIFT $\rightarrow$ UNet. We replace the EIFT module with
the UNet~\cite{OlafRonneberger2015UNetCN} which has a similar
model size with EIFT in the event-image fusion. The two feature maps, $F^{i}_{E}$ and $F^{i}_{I}$, $i = 0,1$, are concatenated and used as the input to the UNet.
2) W/o CCT. We remove the cross-channel transform operation in EIFT.
3) W/o EWP. We remove the element-wise product operation in EIFT.
The PSNR value of these models drops 0.27dB, 0.35dB and 0.29dB respectively.
This result shows the effectiveness of our EIFT module.

\textbf{EGDB}. In the low-light enhancement stage, we also build three modified models for ablation study. 
1) W/o Event Guidance, which means that the map generation block is removed and the input image features are not masked.
2) W/o Global Branch, which removes the global branch in EGDB. Note that we increase the number of the residual blocks in `W/o Global Branch' to make the network parameter amount equal to the full model.
3) W/o Local Branch, which removes the local branch in this model. 
The PSNR value of these models drop 0.32dB, 0.47dB, and 0.39dB respectively.
The result shows the effectiveness of our EGDB module.

\subsubsection{Effectiveness of the Event Guidance.}
To show the effectiveness of the event guidance, we build a model named `PCD+EGDB (W/o EG)'  in Table~\ref{tab:abstudy}.
The input of this model is five consecutive frames, and the network size is similar to the full model.
The first part of `PCD+EGDB (W/o EG)' is the PCD module in~\cite{RuixingWang2021SeeingDS} and the EGDB module of the second part is without the events (the same as W/o Event Guidance).
We can see that the PSNR decreases sharply by 1.54dB, which shows the effectiveness of the event guidance and our pipeline.

\begin{figure}[!t]
\centering
    \includegraphics[width=0.46\textwidth]{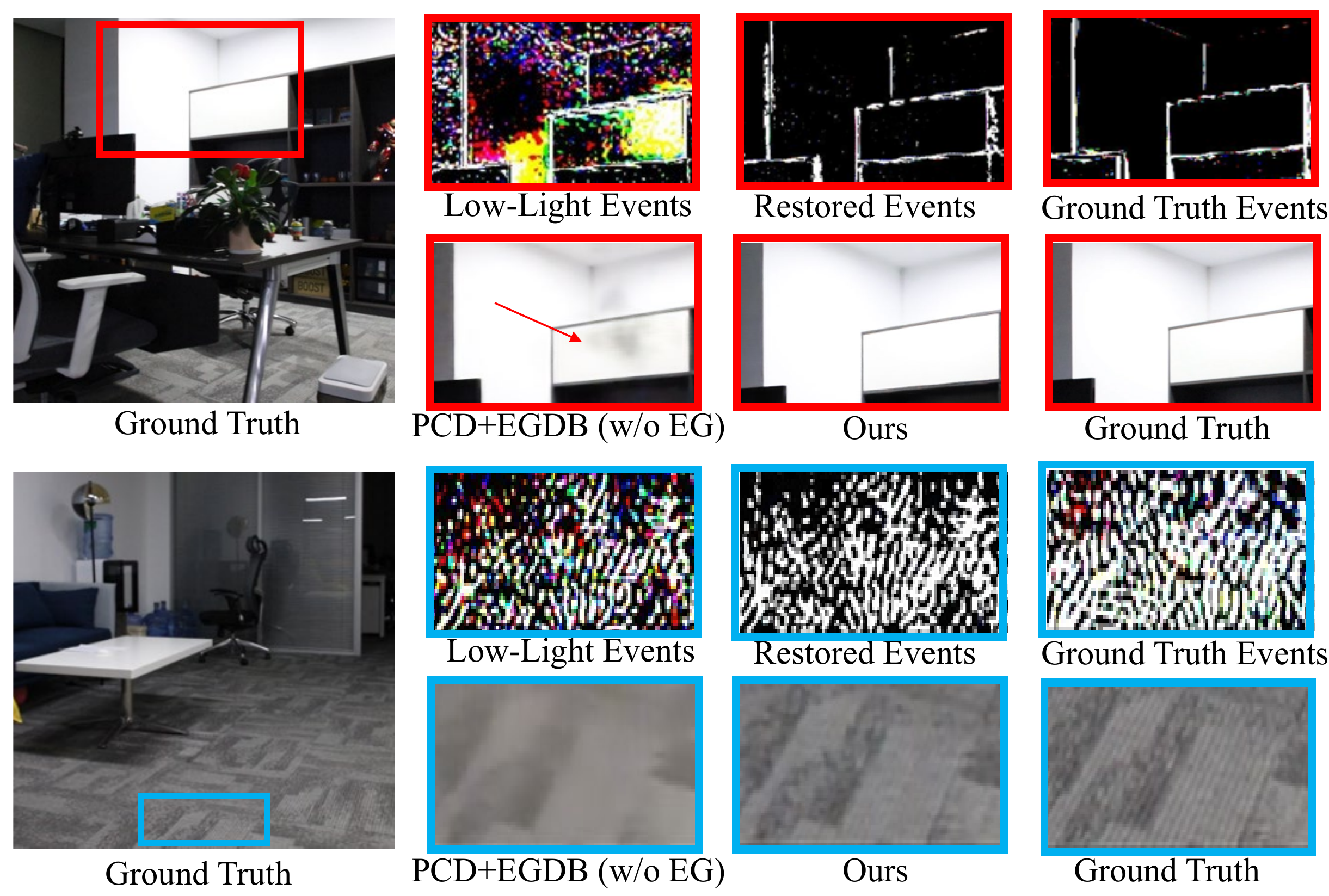}
   \caption{Illustration of how the synthetic events help the restoration of low-light videos. EG: event guidance.}  
     \label{fig:eventsillust}
\end{figure}

\subsection{Visualization of the Restored Event Voxels.}
In Fig.~\ref{fig:eventsillust}, we give some results of restored event voxels to show how the restored events help to enhance the low-light videos.
On the low-light voxels in the top row, we can see that the events contain noise and color shift.
Our event restoration network successfully removes these artifacts.
Our result has much fewer artifacts compared with the prediction of `PCD+EGDB (w/o EG)'.
The bottom row example demonstrates that with less noise in the restored events, our result is sharper and keeps better details.

\subsection{Generalization to Other Real Videos}
 \begin{figure}[!t]
		\centering
		\includegraphics[ width=0.46\textwidth]{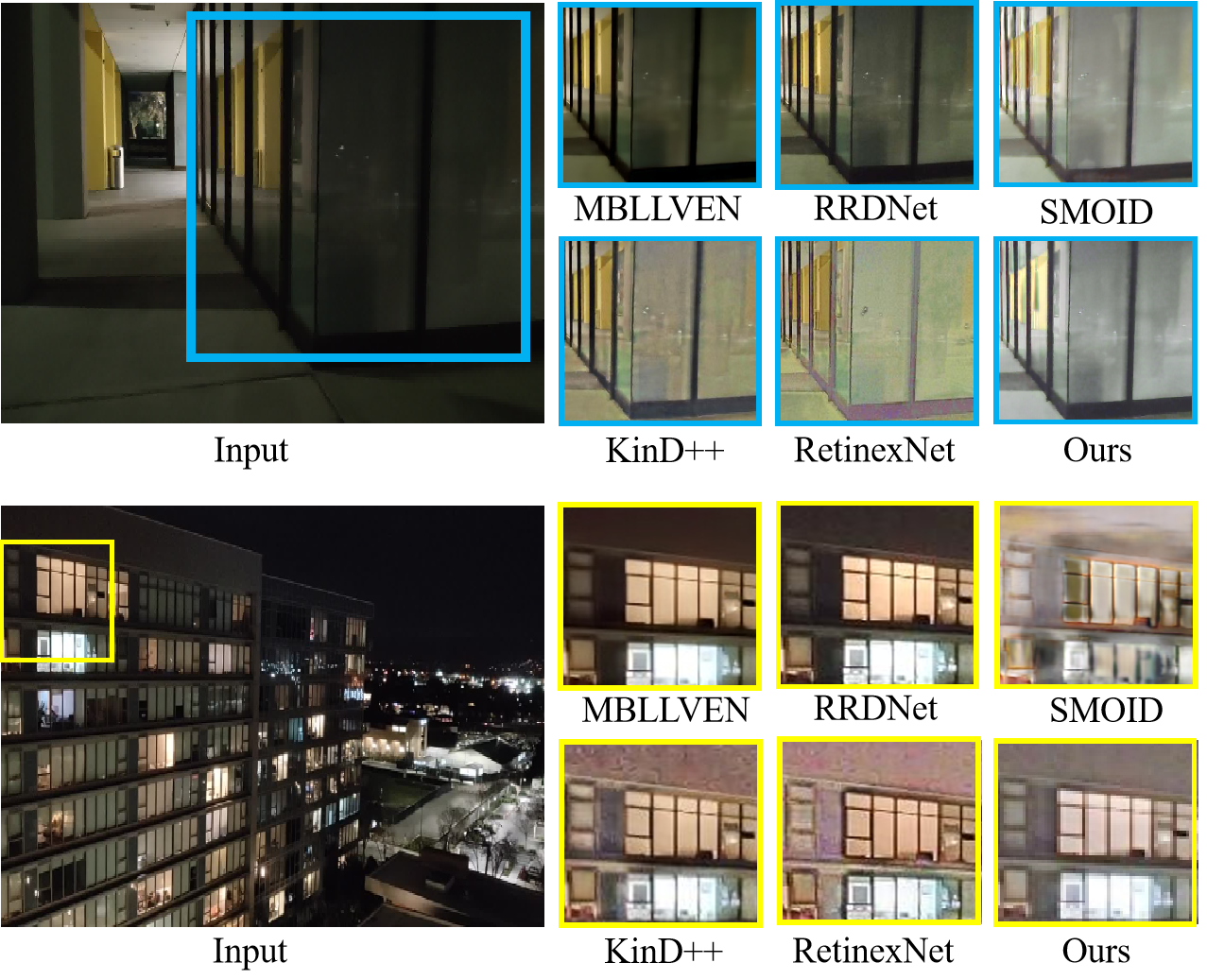}
		\caption{Visual comparison between some SOTA methods and ours on the Loli-Phone dataset.}  
		\label{fig:resultloli}
\end{figure}

To verify the generalization ability of our method on other real low-light videos, we also test it and some SOTA low-light enhancement methods (including MBLLVEN, SMOID, KinD++~\cite{YonghuaZhang2021BeyondBL}, RetinexNet~\cite{ChenWei2018DeepRD}, and RRDNet~\cite{AnqiZhu2020ZeroShotRO}) on the Loli-Phone dataset (without ground truth)~\cite{LoLi}.
Two examples are shown in Fig.~\ref{fig:resultloli}. From the images, we can see that the enhancement of low-light videos by MBLLVEN and RRDNet is limited. 
KinD++ and RetinexNet produce some color artifacts.
The results of SMOID contain multi-frame fusion artifacts (zooming in to see the upper left part of the first result by SMOID). 
The prediction by our method does not have the issues of those methods.
Another advantage of our method is that with the help of events, it better deals with the white balance problem in low light.
Under natural light, the floor and wall should appear white or gray. However, the results predicted by other methods are yellowish, while our results are gray.

\section{Conclusions}
We propose to use synthetic events as the guidance for low-light video enhancement.
We design a novel Event and Image Fusion Transform (EIFT) module for event-image fusion and an Event-Guided Dual Branch (EGDB) module for low-light enhancement.
Our method outperforms both the state-of-the-art low-light video and single image enhancement approaches.
It takes extra time to synthesize the events (about 20$\%$ of the total inference time).
Future work includes exploring better ways to restore the events in low-light conditions and fusing events and images. 

\bibliography{aaai23}

\appendix
\newpage
\section{The Structure of the Global Branch of EGDB}
\begin{figure}[t!]
		\includegraphics[ width=0.49\textwidth]{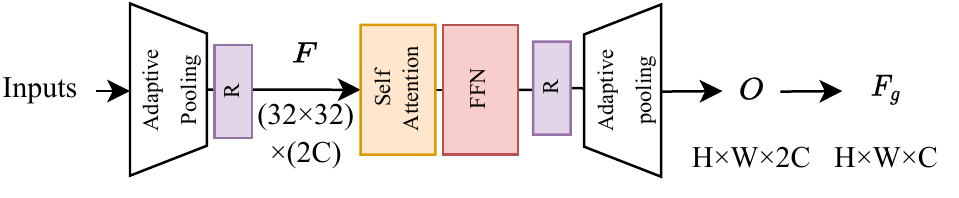}
        
		\caption{Architecture of the global branch of EGDB.} 
		\label{fig:enhancement}
\end{figure}
\begin{figure*}[hb]
		\includegraphics[ width=0.95\textwidth]{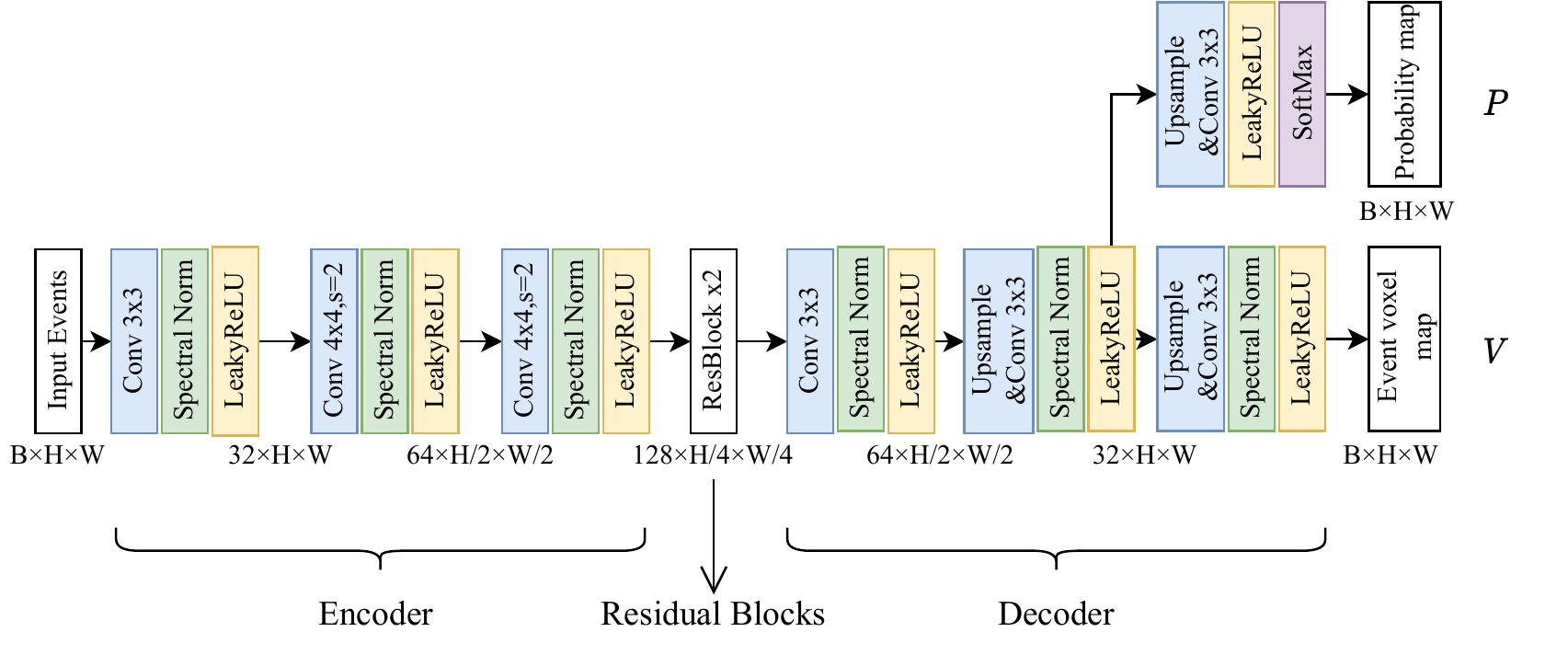}
        
		\caption{Architecture of our proposed event enhancement network. All ReLU layers have been omitted for clarity.} 
		\label{fig:eenet}
\end{figure*}

As shown in Fig.\ref{fig:enhancement}, the inputs are first downsampled through adaptive pooling and then reshaped into $F \in \mathbb{R}^{ 32\times32 \times 2C}$.
The feature $F$ is then splitted into $m$ feature patches $F_{i} \in \mathbb{R}^{ p ^{2} \times 2C}, i=\{1, \ldots, m\}$, where $p$ is the patch size and $m=\frac{H}{p} \times \frac{W}{p}$, and passes a self-attention layer and a feed-forward layer of the conventional Transformer.
The operations of these two layers are represented as:
\begin{equation}
F_{i}^{\prime}=(\mathrm{MHA}\left(\mathrm{NL}(F_{i}), \mathrm{NL}(F_{i}), \mathrm{NL}(F_{i})\right) + F_{i},\\
\end{equation}
\begin{equation}
O_{i}=\mathrm{FFN}\left(\mathrm{NL}(F_{i}^{\prime})\right)+F_{i}^{\prime},
\end{equation}
where $\mathrm{MHA}$, $\mathrm{NL}$, and $\mathrm{FFN}$ represent multi-head attention, layer normalization, and feed-forward network respectively.
The output $O_{1},O_{2},...,O_{m}$ is then reshaped and upsampled using adaptive pooling to $O \in \mathbb{R}^{H \times W \times 2C}$, the first $C$ channels of which is treated as the output $F_{g} \in \mathbb{R}^{H \times W \times C}$ of the global branch.

\section{The Structure of the Event Restoration Network}
Fig.~\ref{fig:eenet} shows the structure of the event restoration network, which contains an encoder, two residual blocks, and a decoder. The encoder downsamples the input events to a feature of one-quarter of the original resolution. The decoder upsamples the feature to the original resolution. 
The network predicts an event probability map $ P \in \mathbb{R}^{B \times H \times W}$ and an event voxel map $V\in \mathbb{R}^{B \times H \times W} $ simultaneously.

\begin{figure*}[!t]
   \centering
		\includegraphics[ width=0.85\textwidth]{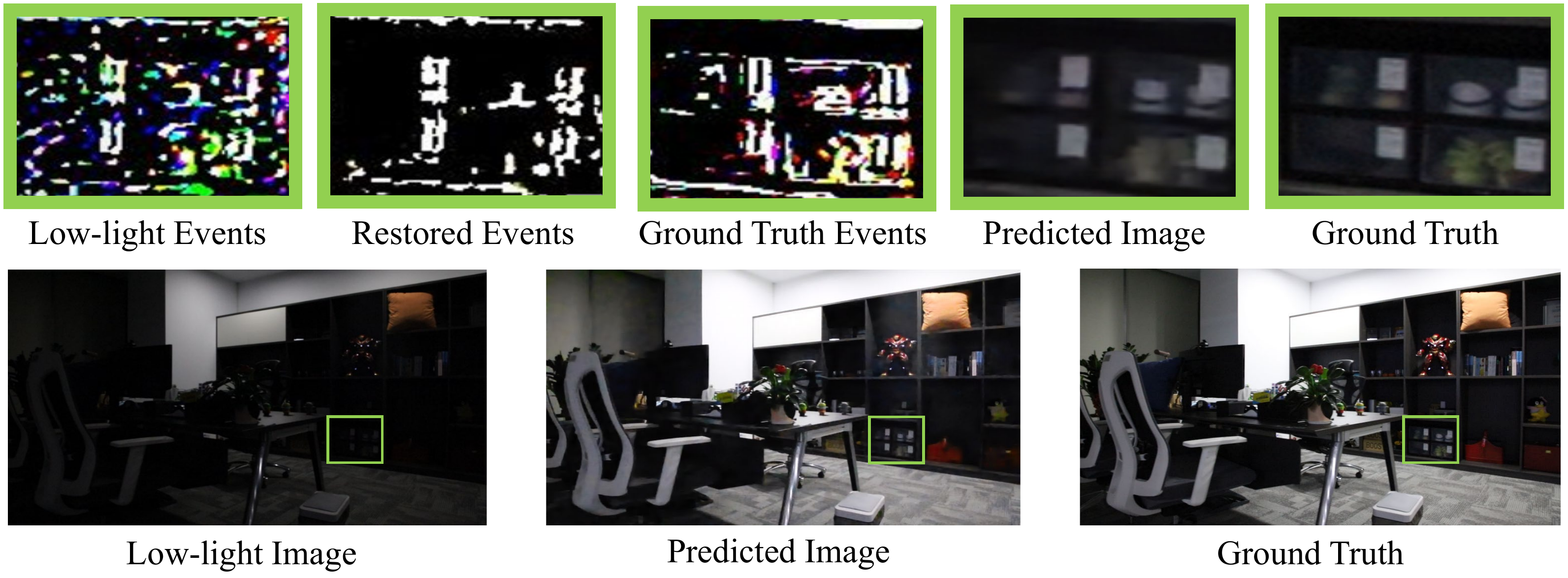}
        
		\caption{The Limitation of our method. When the events are seriously damaged in low-light condition, and the event restoration network can restore some of events, which leads to loss of some small details in the reconstructed RGB image.} 
		\label{fig:limitation}
\end{figure*}

\begin{figure}[!t]
		\centering
		
		\includegraphics[ width=0.48\textwidth]{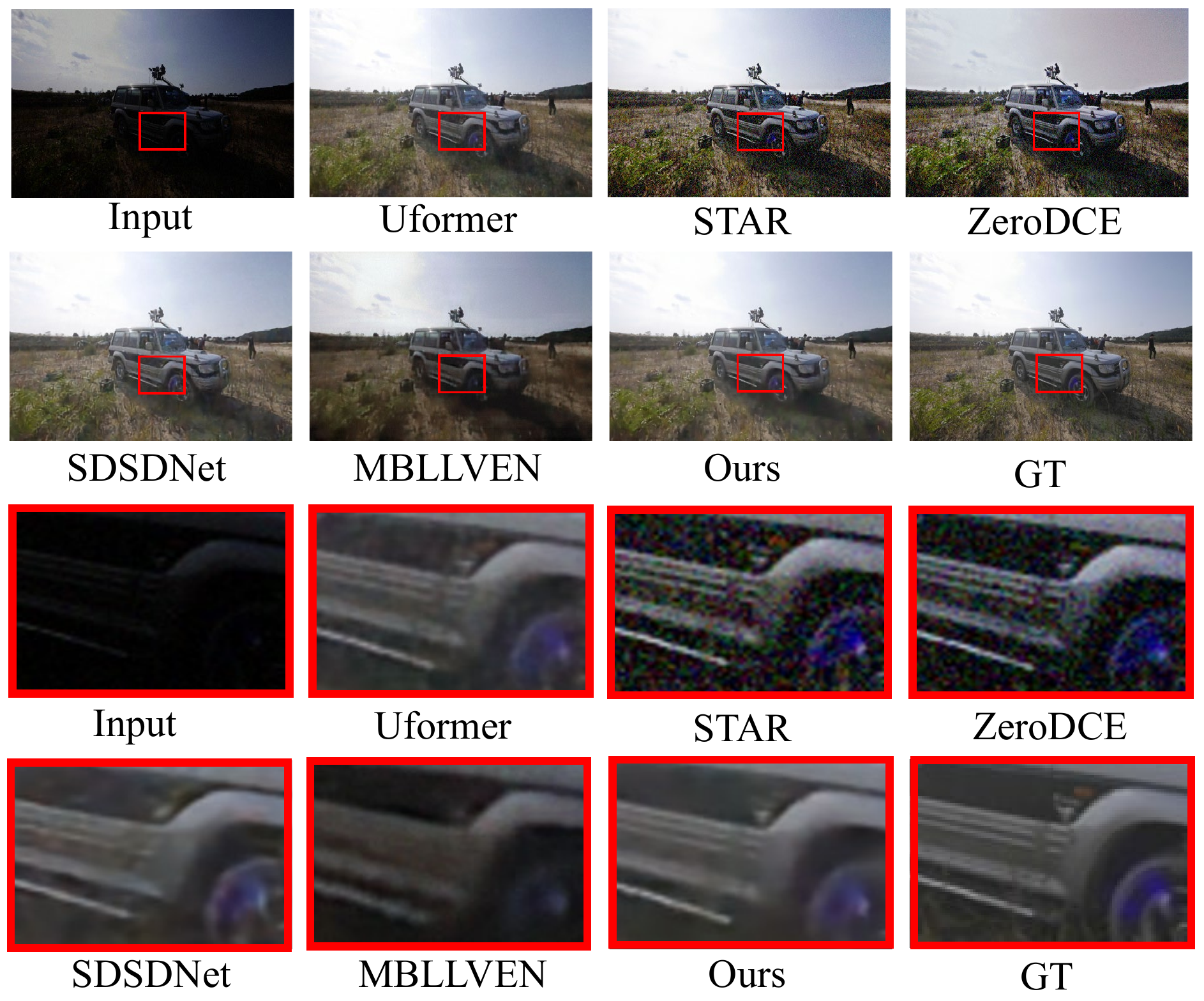}
		\caption{Visual comparison results of some SOTA low-light enhancement methods and our proposed method.}  
		\label{fig:vimeo90result}
\end{figure}

\section{Visual Results on Vimeo90K }

In Fig.~\ref{fig:vimeo90result}, we show qualitative comparison with state-of-the-art methods on Vimeo90K dataset.
The compared methods are Uformer~\cite{ZhendongWang2021UformerAG}, SDSDNet~\cite{RuixingWang2021SeeingDS}, STAR~\cite{ZhaoyangZhang2021STARAS}, ZeroDCE~\cite{ChunleGuo2020ZeroReferenceDC}, and MBLLVEN~\cite{FeifanLv2018MBLLENLI}. 
The single-image-based methods produce some noise (Uformer, ZeroDCE, and STAR).  
As for video-based methods, the illumination estimation of MBLLVEN is insufficient, which leads to darker result.
SDSDNet produces some noise artifacts and loss of details.
Our method achieves the best results through restoring the color and removing the noise simultaneously with the help of events.

\section{The Process of Synthesizing Low-Light Videos on Vimeo90K}

In the experiments, we synthesize a large-scale data from the Vimeo90K dataset. %
We select video clips whose average brightness is greater than 0.3 (the highest brightness is 1) as normal-light ground truth, and then use the method in ~\cite{lv2021attention,FanZhang2022LearningTC} to synthesize low light video sequences.
The ground truth videos are darkened using gamma correction and linear scaling, which can be formulated as:
 \begin{equation}
S_{l}=\beta \times(\alpha \times S_{g})^{\gamma} + N_{\sigma}\text {, }
\end{equation}
where $S_{l}$ and $S_{g}$ are low-light sequence and ground truth sequence, respectively;  $\gamma$, $\alpha$, and $\beta$ are gamma correction, normal-light linear scaling factor, and low-light linear scaling factor respectively; $N_{\sigma}$ is the Gaussian noise map with standard deviation $\sigma$.
In the training stage, $\gamma$, $\alpha$, $\beta$, and $\sigma$ are sampled from uniform distributions $U (2, 3.5)$, $U(0.9, 1)$, $U(0.5, 1)$, and $U(0, 0.02)$, respectively.
In the testing stage, $\gamma$, $\alpha$, $\beta$, and $\sigma$ are set to 2.75, 0.95, 0.8 and 0.01 respectively.

\section{The Limitation}
In this section, we show the limitation of our method.
In Fig.~\ref{fig:limitation}, the low-light events are seriously damaged, which prevents us from restoring good events.
Compared with the ground truth, the final restored image loses some details.
Our future will investigate on how to restore the low-light events better.

\section{The Difference between Event Information and Edge Information.}
Although the processed continuous event voxels are similar to edges extracted directly from images, they still have many differences. 
1) Event voxels are transferred from discrete events, which are not continuous themselves.
Discrete events contain temporal information, while edges from a single image do not have temporal information.
2) Event voxels contain time-varying motion information of videos.
As shown in Fig.~\ref{fig:limitation}, the intensity of the horizontal and vertical edges of the four white rectangles in Ground Truth should be the same in edge extraction.
However, the camera capturing the video is moving horizontally, so in the event voxels, the vertical events are thicker or more responsive than the horizontal events.
\end{document}